\documentclass[10pt,twocolumn,letterpaper]{article}

\usepackage{iccv}
\usepackage{amsmath}
\usepackage{amssymb}
\usepackage{epsfig}
\usepackage{graphicx}
\usepackage{times}

\usepackage{booktabs}

\usepackage{makecell}
\usepackage{multicol}
\usepackage{multirow}
\usepackage{tabularx}
\usepackage[dvipsnames]{xcolor}
\usepackage{subcaption}

\usepackage{pifont}
\newcommand{\cmark}{\ding{51}}%
\newcommand{\xmark}{\ding{55}}%

\usepackage{calc}
\newlength\myheight
\newlength\mydepth
\settototalheight\myheight{Xygp}
\newcommand*\inlinegraphics[1]{%
  \settototalheight\myheight{Xygp}%
  \settodepth\mydepth{Xygp}%
  \raisebox{-\mydepth}{\includegraphics[height=\myheight]{#1}}%
}

\usepackage[pagebackref=true,breaklinks=true,colorlinks,bookmarks=false]{hyperref}

\newcommand{\kevin}[1]{\textcolor{black}{#1}}

\usepackage{amsmath}
\usepackage{etoolbox}
\usepackage{bm}
\usepackage{amsthm}
\usepackage{mathtools}
\usepackage{algorithm}
\usepackage{algpseudocode}
\usepackage{xspace}

\newcommand{\MyMapTemplateNoPrefix}[3]{\expandafter#1\csname#3\endcsname{#2{#3}}}
\forcsvlist{\MyMapTemplateNoPrefix {\def} {\mathbf} } {0,1,a,b,c,d,e, f, g, h, i, j, k, l, m, n, o, p, q, r, u, v, w, x, y, z} 

\newcommand{\MyMapTemplatePrefix}[4]{\expandafter#1\csname#3#4\endcsname{#2{#4}}} 
\forcsvlist{\MyMapTemplatePrefix {\def} {\mathcal}{c}} {A,B,C,D,E,F,G,H,I,J,K,L,M,N,O,P,Q,R,S,T,U,V,W,X,Y,Z}  
\forcsvlist{\MyMapTemplatePrefix {\def} {\mathbf} {b}} {A,B,C,D,E,F,G,H,I,J,K,L,M,N,O,P,Q,R,S,T,U,V,W,X,Y,Z,RP} 
\forcsvlist{\MyMapTemplatePrefix {\DeclareMathOperator} {} {} } {trace, argmax, argmin, RoIPool, sign}

\def\all{{\mathrm{all}}}

\iccvfinalcopy 

\def\iccvPaperID{****} 

\ificcvfinal\fi
\begin{document}

\title{\vspace{-15pt} OVIS: Open-Vocabulary Visual Instance Search via \\ Visual-Semantic Aligned Representation Learning\vspace{-5pt}} 

\author{Sheng Liu$^{1}$ \qquad Kevin Lin$^{2}$ \qquad Lijuan Wang$^{2}$ \qquad Junsong Yuan$^{1}$ \qquad Zicheng Liu$^{2}$\\
 $^{1}$University at Buffalo \qquad  $^{2}$Microsoft\\
{\tt\small \{sliu66, jsyuan\}@buffalo.edu \qquad \{keli, lijuanw, zliu\}@microsoft.com}
}

\twocolumn[{%
\renewcommand\twocolumn[1][]{#1}%
\maketitle
\begin{center}
    \centering
    \vspace{-13pt}
    \includegraphics[width=.9\linewidth]{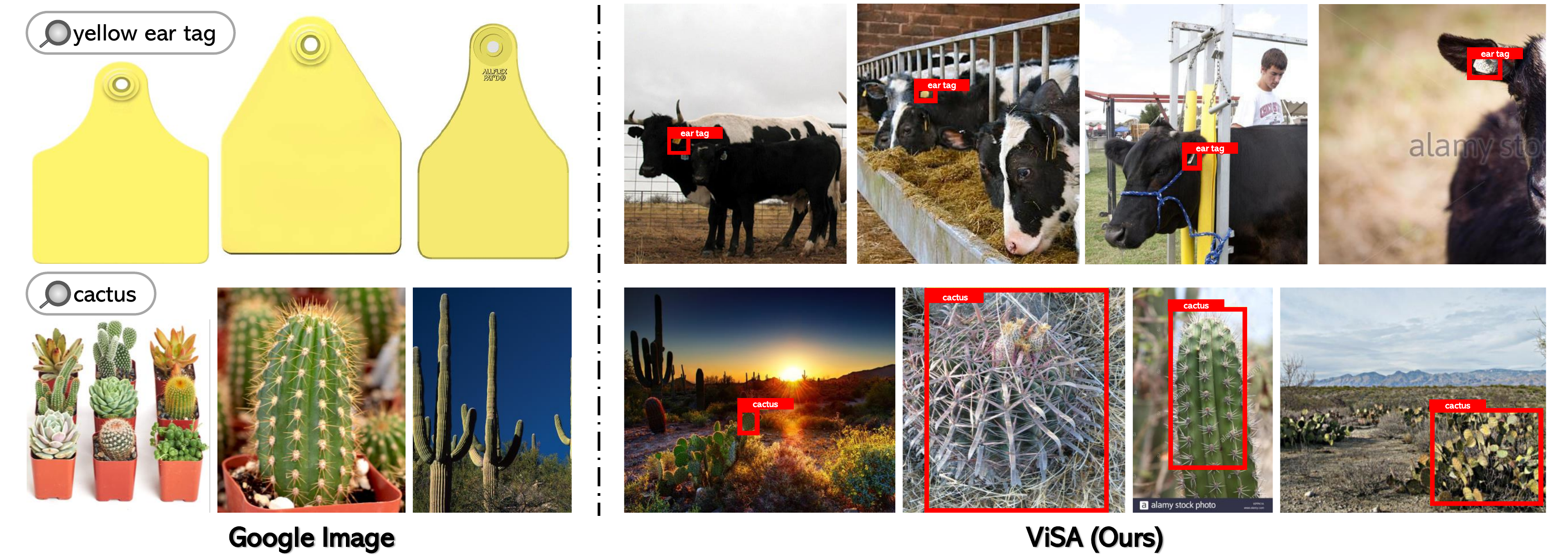}
    \vspace{-5pt}
    \captionof{figure}{Comparison between Google Image and ViSA, \textit{i.e.}, our model for OVIS. Similarities: both take as input a textual search query, \textit{e.g.}, \texttt{yellow ear tag}, \texttt{cactus}, and return a ranked list of visual contents. Differences: (1) Google Image returns images, while ViSA returns visual instances, \textit{i.e.}, localized regions (shown with red bounding boxes). ViSA is able to find \textit{small} visual instances and visual instances in \textit{context}. (2) Google Image heavily relies on textual metadata of images, while ViSA relies on visual contents only (more details are provided in the supplementary materials).}
    \label{fig:figures}
\end{center}%
}]

\maketitle
\ificcvfinal\thispagestyle{empty}\fi

\begin{abstract}
We introduce the task of open-vocabulary visual instance search (OVIS).
Given an arbitrary textual search query, Open-vocabulary Visual Instance Search (OVIS) aims to return a ranked list of visual instances, \textit{i.e}., image patches, that satisfies the search intent from an image database. The term ``open vocabulary'' means that there are neither restrictions to the visual instance to be searched nor restrictions to the word that can be used to compose the textual search query. We propose to address such a search challenge via visual-semantic aligned representation learning (ViSA). ViSA leverages massive image-caption pairs as weak image-level (not instance-level) supervision to learn a rich cross-modal semantic space where the representations of visual instances (not images) and those of textual queries are aligned, thus allowing us to measure the similarities between any visual instance and an arbitrary textual query. To evaluate the performance of ViSA, we build two datasets named OVIS40 and OVIS1600 and also introduce a pipeline for error analysis. Through extensive experiments on the two datasets, we demonstrate ViSA's ability to search for visual instances in images not available during training given a wide range of textual queries including those composed of uncommon words. Experimental results show that ViSA achieves an mAP$@$50 of 21.9$\%$ on OVIS40 under the most challenging setting and achieves an mAP$@$6 of 14.9$\%$ on OVIS1600 dataset.
\end{abstract}

\section{Introduction}

The sheer number of image searches perfectly reflects its importance. Tens of millions of image searches are carried out in a single day by image search engines, \textit{e.g.}, Google \cite{google}, in a single day. Taking a textual search query, \textit{e.g.}, a word ``\texttt{ovis}''\footnote{\texttt{ovis}: a genus of mammals, whose species are known as sheep \cite{ovis}.} as input, an image search engine returns a list of images relevant to the query. In this sense, an image search engine can be viewed as mapping textual search queries to visual search results. Despite promising text-to-image search results, image search engines like Google often rely on textual descriptions of images, \textit{e.g.}, alt-texts and titles, and not on visual contents of images. In addition, existing image search engine typically returns a whole image rather than locating the textual query in the image. 

In this work, we introduce the task of open-vocabulary visual instance search (OVIS). Given a textual search query, \textit{e.g.}, ``\texttt{ovis}'', ``\texttt{marble column}'', OVIS aims to return visual instances, \textit{i.e.}, image patches (instead of images)\footnote{The terms, ``visual instance'' and ``image patch'', are used interchangeably in this manuscript.}, which are relevant to the query, solely relying on the visual contents of images. We use the term ``open'' as we do not limit the visual instances that can be searched, it can be instances of any objects, movements and attributes. In contrast, works on image retrieval mainly focus on retrieving \textit{whole images} of a \textit{closed} set of classes \cite{liu2016deep,cao2018deep,yu2017hope,yu2018product,yu2020product, johnson2015image, faghri2017vse++, zhang2020context}. Furthermore, we do not restrict the words that can be used in the textual queries. Words from any part of speech can be used, \textit{e.g.}, nouns, verbs and adjectives.

The vast number of the visual instances to be searched and the textual search queries makes OVIS a challenge. While state-of-the-art computer vision models have achieved great success in many areas, they often have a closed vocabulary limited by the annotated categories. The vocabulary of an object detector is limited, for example, by the number of object classes with bounding box annotations. They cannot detect classes of objects with no bounding box annotations. However, it is infeasible for us to create a sufficiently large dataset that, covers all the possibilities of the visual instances as well as the textual search queries, due to their large numbers.

To address this challenge, we propose to use a large number of image captions that can be collected by a web crawler to train our model. However, captions describe images rather than visual instances. Therefore, captions can only serve as weak supervision, as we have to associate words / phrases of the captions with visual instances in images without explicit supervision. This is achieved with the help of masked token prediction, which is a task that attempts to predict the masked token in the caption based on visual instances in the image and the other tokens. In order to correctly predict the masked token, our model must attend to visual instances relevant to the masked token. In this way, an implicit association is achieved. As a result, our model is able to encode visual instances and textual search queries into representations that are aligned in a common semantic space. In other words, visual instances and textual search queries with similar semantics have similar representations. We also use a small number of textual visual instance labels so that our model can explicitly associate visual instances to tokens in the labels during training. While we only use a \textit{small closed} set of textual labels, they serve as anchors that ease the learning of the alignment between the representations of visual instances and textual queries.

We collect OVIS40 and OVIS1600 datasets with $\sim6$K and $\sim5$K visual instances, which corresponds to $40$ and $1,600$ sophisticated queries with different characteristics in order to evaluate our model. These two datasets can serve as benchmarks for future research in this direction. In addition, we propose an error analysis pipeline with which the sources of error in OVIS models can be analyzed.

\renewcommand{\arraystretch}{1.0}
\setlength{\tabcolsep}{2pt}
\begin{table}[t]
    \centering
    \begin{tabular}{ccccc}
    \toprule
     & IR & WSOD & OV-CLS & OVIS \\
    \midrule
    Incomplete supervision? & \xmark & \cmark & \cmark & \cmark \\
   \midrule
    Instance-level? & \xmark & {\color{OrangeRed}\cmark} & \xmark & {\color{OrangeRed}\cmark} \\
    \midrule
    Open-vocabulary? & \xmark & \xmark & {\color{OrangeRed}\cmark$^{*}$} & {\color{OrangeRed}\cmark} \\
    \bottomrule
    \end{tabular}
    \vspace{-2pt}
    \caption{Comparison of different tasks related to OVIS. IR: image retrieval; WSOD: weakly supervised object detection; OV-CLS: open-vocabulary image classification \cite{frome2013devise} ($^{*}$ indicates that \cite{frome2013devise} is not able to classify images whose labels do not have word vectors).}
    \vspace{-15pt}
    \label{tab:tasks}
\end{table}

\section{Backgrounds} 

We compare OVIS with three related tasks: image retrieval (IR), weakly supervised object detection (WSOD) and open-vocabulary image classification (OV-CLS). Key features of these tasks are shown in Table \ref{tab:tasks}. In addition to these tasks, OVIS is also \kevin{closely} related to image-text retrieval and large-vocabulary instance segmentation \cite{gupta2019lvis, wu2020forest}.

\noindent \textbf{Image Retrieval (IR):} Given an image as input, the goal of IR is to retrieve images that are similar or have similar semantics to the given image in an image database. Supervised hashing \cite{liu2016deep,cao2018deep,yu2017hope,yu2018product,yu2020product} has become a paradigm for IR due to its low computational cost. In contrast to OVIS, IR models are trained to retrieve images of a \textit{closed} set of classes in a supervised manner. Moreover, IR models retrieve images, while OVIS retrieves visual instances.

\noindent \textbf{Weakly Supervised Object Detection (WSOD):} WSOD aims to train an object detector without using bounding-box annotations. \kevin{Prior studies}~\cite{ren2020instance, huang2020comprehensive, zeng2019wsod2, yang2019towards, wen2016discriminative, hong2017fried, hong2019asymmetric} use image tags as supervision, and Ye~\textit{et al.}~\cite{ye2019cap2det} use image-caption pairs as supervision. \kevin{In addition, Weakly Supervised Object Localization (WSOL)~\cite{zhou2016learning,bilen2016weakly,choe2020evaluating,oquab2015object} is a related topic to WSOD. WSOL aims to localize a single class-specific region in an image.} Contrary to OVIS, WSOD \kevin{and WSOL only focus} on a fixed set of object classes. 

\noindent \textbf{Open-Vocabulary Image Classification (OV-CLS):} Given an image as input, the goal of OV-CLS \cite{frome2013devise} is to assign a class label to the image. The main difference between OVIS and OV-CLS is that OV-CLS assigns image-level labels, while OVIS returns a list of visual instances, \textit{i.e.}, image patches.

\noindent \textbf{Vision-Language Pre-Training:} Vision-language pre-trained (VLP) models \cite{li2020oscar, chen2020uniter, zhou2020unified, su2019vl, tan2019lxmert, li2020hero, cao2020behind, li2020weakly} are highly successful in learning cross-modal representations for \kevin{various vision-language tasks using image captions as supervision}. 
Our model is not a VLP model, as our model is directly applied to OVIS after being trained, while VLP models have to be \textit{finetuned} for \textit{downstream} tasks. In addition, our model is mainly trained in a weakly-supervised manner as captions only provide image-level (not instance-level) annotations, while VLP models have to be finetuned in a supervised manner.

\begin{figure}
    \centering
    \includegraphics[trim={0, 0, 20pt, 25pt}, clip, width=0.9\linewidth]{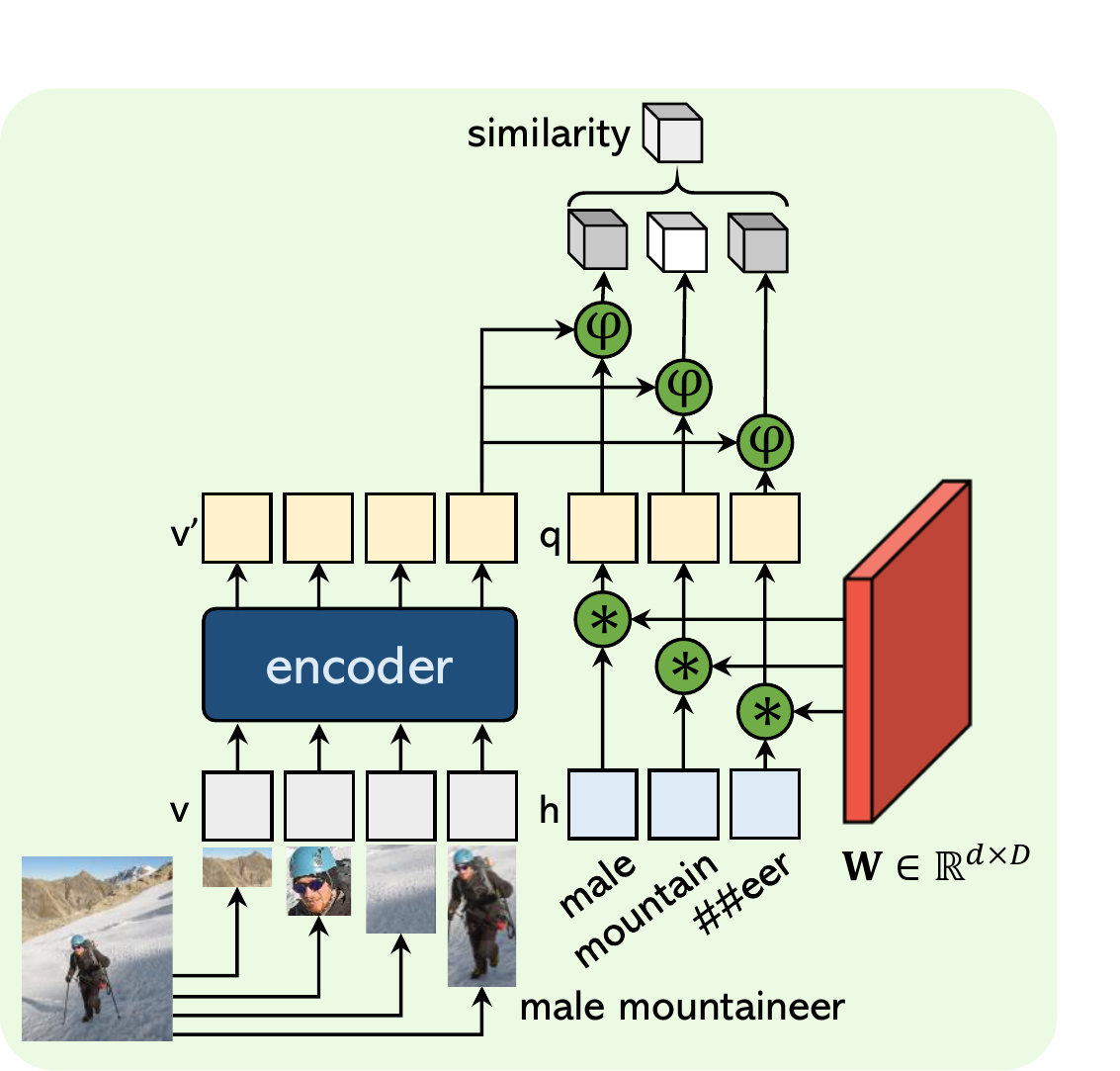}
    \caption{The way our model computes the similarity between a textual query ``\texttt{male mountaineer}'' and the $4$-th visual instance in an image at test time. Our model consists of a visual-semantic encoder and a base-token embedding matrix $\mathbf{W}$. \protect\inlinegraphics{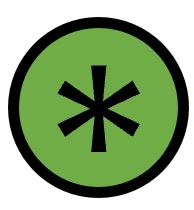}: matrix multiplication operation; \protect\inlinegraphics{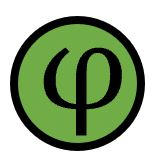}: similarity measure, \textit{e.g.}, cosine similarity.}
    \vspace{-15pt}
    \label{fig:test}
\end{figure}

\section{Method}
\label{sec:method}

In this section, we first introduce how our model can be used at test time (assuming it has been trained). We then introduce how we train our model via visual-semantic aligned representation learning. To this end, we discuss a preprocessing scheme that could be used to speed up the search process.

\noindent \textbf{Inference:} Essentially, a search problem, \textit{e.g.}, OVIS, can be solved once we are able to measure the similarity between a search query and the items to be searched in a database, as items can be ranked and selected according to their similarity with the given query. In our case, we aim to compute the similarity between a textual search query, \textit{i.e.}, an \textit{arbitrary} word or phrase consisting of less than $4$ words\footnote{We focus on short queries as more than $80\%$ of web search queries have less than $4$ words \cite{spink2001searching}.} in the set of all $\mathit{147}$\textit{K} words in current use, and an \textit{arbitrary} visual instance. 

As shown in Figure \ref{fig:test}, our model consists of a visual-semantic encoder, \textit{i.e.}, a Transformer encoder \cite{vaswani2017attention}, and a base-token embedding matrix $\mathbf{W} \in \mathbb{R}^{d \times D}$ \inlinegraphics{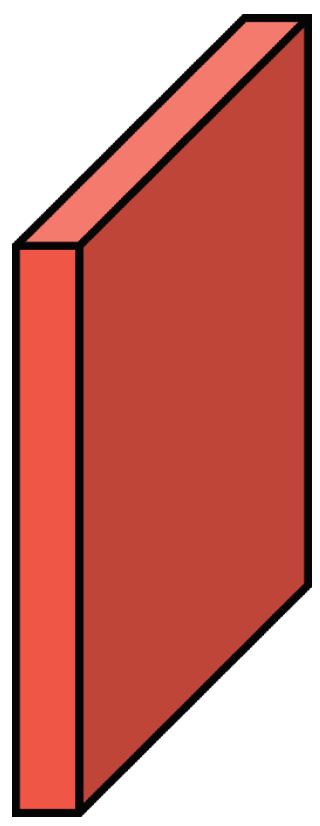}, where $D$ is the size of the dictionary of our model. 

Given a textual query, we tokenize the query into a set of tokens in the dictionary of our model. For example, ``\texttt{male mountaineer}'' is tokenized into ``\texttt{male}'', ``\texttt{mountain}'' and ``\texttt{\#\#eer}''. Thanks to tokenization, our model can handle any word in the set of $147$\textit{K} words in current use, even if it does not appear in our model's dictionary, \textit{e.g.}, ``\texttt{mountaineer}''. We then encode the tokens into vector representations in a semantic space $\mathcal{S}$ via $\q_i = \mathbf{W} \boldsymbol{\cdot} \h_i$, where $\q_i \in \mathbb{R}^{d}$ and $\h_i \in \{0, 1\}^{D}$ denote the vector representation \inlinegraphics{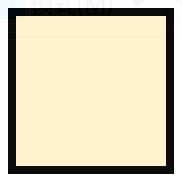} and one-hot vector \inlinegraphics{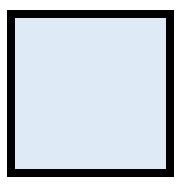} of the $i$-th token. In other words, we encode each token using a column of the base-token embedding matrix $\mathbf{W}$. 

Given an image $\mathbf{I}$, we use a pretrained visual backbone to identify $n$ visual instances in it and extract their features \inlinegraphics{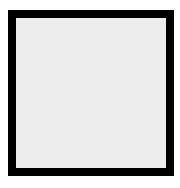}, $[\v_1, \v_2, \ldots, \v_n], \v_j \in \mathbf{R}^{d}$. The sequence of features are encoded \textit{jointly} by our visual-semantic encoder into a sequence of \textit{contextualized} representations \inlinegraphics{figures/v.jpg}, $[\v^{'}_1, \v^{'}_2, \ldots, \v^{'}_n], \v^{'}_j \in \mathbf{R}^{d}$, which are in the same semantic space $\mathcal{S}$ as token representations. 

We then compute the similarity between the representation of a visual instance $\v^{'}_j$ and the representation of each token $\q_i$ with a similarity measure $\psi: \mathbb{R}^{d} \times \mathbb{R}^{d} \rightarrow \mathbb{R}$, \textit{e.g.}, cosine similarity. We will compare different instantiation of $\psi$ in Section \ref{sec:ovis40-exp}. The similarity between a visual instance and a textual query is the average of the similarity between the visual instance's representation and each token's representation computed with $\psi$. Visual instances are ranked according to their similarities with the textual query. 

To ensure that the computed similarities are meaningful, it is essential that the representations of both the tokens, \textit{i.e.}, columns of $\mathbf{W}$, and those of visual instances are aligned in the same semantic space $\mathcal{S}$. Similarity between representations in different semantic space is not meaningful, for example, similarity between the feature of a visual instance $\v_j$ and the one-hot vector of a token $\h_i$ is meaningless. Therefore, our goal is to train the visual-semantic encoder and the base-token embedding matrix $\mathbf{W}$ so that they can align representations of visual instances and tokens in a common semantic space $\mathcal{S}$. In other words, our goal is to ensure representations of visual instances and tokens with similar semantics have great similarities, while those with different semantics have little similarities, for example, visual instances of a mountaineer are very similar to token ``\texttt{mountain}'' and token ``\texttt{\#\#eer}'' and have little similarities with token ``\texttt{dolphin}''.

\begin{figure}
    \centering
    \includegraphics[trim={0, 0pt, 35pt, 10pt}, clip, width=0.85\linewidth]{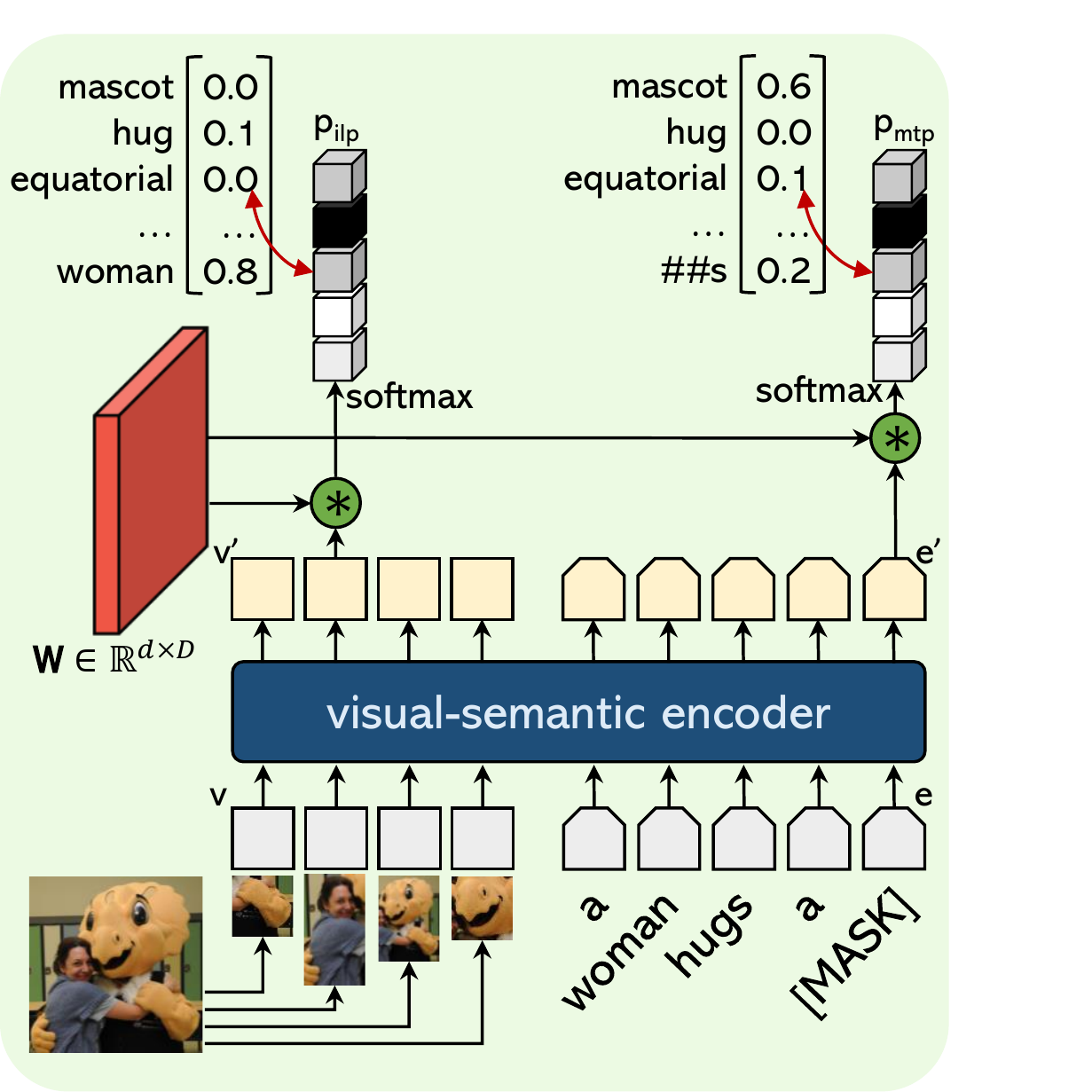}
    \caption{The way our model is trained. Each training sample is an image-caption pair, represented as visual instance features \protect\inlinegraphics{figures/v0.jpg} and token embeddings \protect\inlinegraphics{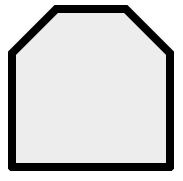}. We train our model using two tasks: masked token prediction (MTP), that aims to predict the masked token, and visual instance label prediction (ILP) whose goal is to predict the textual labels for visual instances that belong to a \textit{small closed} set of classes. \protect\inlinegraphics{figures/mul.jpg}: matrix multiplication operation.}
    \vspace{-15pt}
    \label{fig:learning-v2}
\end{figure}

\noindent \textbf{Visual-Semantic Aligned Representation Learning:} Should we were able to build a dataset containing all possible visual instances and all possible textual search queries with which to search them, we would be able to learn such an alignment in a supervised manner by directly maximizing the similarity between a visual instance and search query whose semantics are alike. However, it is infeasible to build such an enormous dataset. Therefore, we propose to learn the alignment via visual-semantic aligned representation learning, which mainly leverages image captions collected by a web crawler, as image-level supervision. As captions describe images instead of visual instances, it is therefore important that, during training, we can make associations between words / phrases in the captions and visual instances in images.

To achieve such a goal, we simply mask a percentage of tokens (replace with a special ``[\texttt{MASK}]'' token) in a caption at random (\textit{e.g.}, the $5$-th token in Figure \ref{fig:learning-v2}) and then predict the masked token from the other tokens in the caption and the visual instances in the image described by the caption. Such a process is often referred to as masked token prediction (MTP). As shown in  Figure \ref{fig:learning-v2}, the visual-semantic encoder takes a concatenation of $m$ caption token embeddings $[\e_1, \e_2, \ldots, \e_m], \e_j \in \mathbb{R}^{d}$ and $n$ visual instance features $[\x_1, \x_2, \ldots, \x_n], \x_i \in \mathbb{R}^{d}$ as input. It encodes both of them together into $[\e^{'}_1, \e^{'}_2, \ldots, \e^{'}_m]$ and $[\v^{'}_1, \v^{'}_2, \ldots, \v^{'}_n]$. $\e^{'}_{i}$ and $\v^{'}_{j}$ are contextualized representations of $\e_i$ and $\v_j$, respectively. Suppose the $i$-th token is masked (replaced by ``[\texttt{MASK}]''). We then predict the $i$-th token via $\l = \e_{i}^{'} \boldsymbol{\cdot} \mathbf{W}, \l \in \mathbb{R}^{D}$. $\l$ contains logits, which are then normalized into probabilities $\p_{\mathrm{mtp}}$ using $\mathrm{softmax}$ function ($\p_{\mathrm{mtp}}$ is shown on the top right part of Figure \ref{fig:learning-v2}). During training, we adopt a negative log-likelilood (NLL) loss, which allows our model to maximize the probability of the ground-truth masked token, \textit{e.g.}, the probability of ``\texttt{mascot}'' shown as the first element of $\p$ in Figure \ref{fig:learning-v2}.

While MTP enables our model to learn the alignment implicitly with an ``intermediary'', we propose to learn the alignment explicitly without the ``intermediary''. To do this, we let our model predict the textual class labels of visual instances, which belong to a \textit{small closed} set of classes. We refer to this process as visual instance label prediction (ILP). As shown in Figure \ref{fig:learning-v2}, ILP is done similar to MTP. We use NLL loss as the loss function for ILP so that our model can predict the correct textual label for the visual instances, \textit{e.g.}, ``\texttt{woman}'' for the second visual instance in Figure \ref{fig:learning-v2}. Since we only predict labels of visual instances of a closed set of classes, such a loss is only applied to labelled visual instances, for example, it is only applied to the $2$nd one of the class ``\texttt{woman}'' in Figure \ref{fig:learning-v2} as other visual instances are not labelled. If an image does not have any labelled visual instance, we do not apply such a loss at all. While ILP is applied to visual instances of a closed set of classes, the representations of these visual instances \inlinegraphics{figures/v.jpg} are aligned directly with columns of $\mathbf{W}$ without the ``intermediary'', \textit{i.e.}, tokens' representations \inlinegraphics{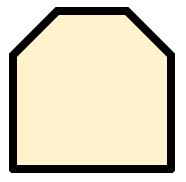}. Hence, the representations of these visual instances can serve as anchors that facilitate learning of representations of other ``open-vocabulary'' visual instances via MTP. In this sense, MTP and ILP complement to each other. We train our model by minimizing the sum of the loss of MTP and that of ILP.

\noindent \textbf{Preprocessing Scheme:} As mentioned when we introduce our inference scheme, the similarity between a visual instance and a textual query is indeed the average of the similarities between the visual instance's representation and columns in the base-token embedding matrix $\mathbf{W}$ which represents tokens in the textual query. Therefore, we can pre-compute and store the similarities between all visual instances in the image database to be searched and all columns in $\mathbf{W}$ (as shown in Figure \ref{fig:test}). Such a process speeds up the search process at test time, because there is no need to compute the similarities at test time (computing similarities between $d$ dimensional vector representations of visual instances and tokens is relatively time-consuming). In practice, indexing methods, \textit{e.g.}, KD-tree \cite{bentley1975multidimensional}, can be used to further accelerate the search process. Fast nearest neighbor methods \cite{berchtold1998fast, hwang2012fast, li2016adaptive, yu2017hope, yu2018product, yu2020product, hong2019asymmetric} can also be used if for some reason there is need to compute similarities at test time. However, that is not the focus of this paper.

\setlength{\tabcolsep}{3pt}
\renewcommand{\arraystretch}{1.0}
\begin{table*}[h]
    \centering
    \begin{tabular}{lcccccccccccc}
    \toprule
    & \multicolumn{4}{c}{OVIS40-small} & \multicolumn{4}{c}{OVIS40-medium} & \multicolumn{4}{c}{OVIS40-large} \\
    \cmidrule(lr){2-5} \cmidrule(lr){6-9} \cmidrule(lr){10-13}
    Model & mAP$_{30}$ & mAP$_{50}$ & mAP$_{70}$ & mAP$_\mathrm{all}$ & mAP$_{30}$ & mAP$_{50}$ & mAP$_{70}$ & mAP$_\mathrm{all}$ & mAP$_{30}$ & mAP$_{50}$ & mAP$_{70}$ & mAP$_\mathrm{all}$ \\
    \midrule
    Det+DeViSE \cite{frome2013devise} & 8.6 & 5.1 & 2.4 & 5.4 & 8.0 & 4.2 & 1.9 & 4.7 & 5.0 & 2.8 & 1.1 & 3.0 \\
    ViSA & {\color{OrangeRed}50.8} & {\color{OrangeRed}35.0} & {\color{OrangeRed}18.5} & {\color{OrangeRed}34.8} & {\color{OrangeRed}46.9} & {\color{OrangeRed}30.6} & {\color{OrangeRed}14.4} & {\color{OrangeRed}30.6} & {\color{OrangeRed}33.6} & {\color{OrangeRed}21.9} & {\color{OrangeRed}10.3} & {\color{OrangeRed}21.9} \\
    \bottomrule
    \end{tabular}
    \vspace{-6pt}
    \caption{Comparison of ViSA and a combination of a detector and DeViSE, \textit{i.e.}, Det+DeViSE, on three subsets of OVIS40.}
    \vspace{-10pt}
    \label{tab:ovis40-devise}
\end{table*}
\setlength{\tabcolsep}{6pt}

\section{Datasets}
\label{sec:dataset}

We create two datasets, \textit{i.e.}, OVIS40 and OVIS1600, to benchmark OVIS methods. Both datasets contain $\sim$117K images, that differ considerably in contents, resolutions, and so on. In order to better simulate a real image database, the two datasets contain not only natural color images, but also man-made images, \textit{e.g.}, cartoons, and grayscale images. There is no overlap between images in the two datasets and those in the training corpus. 

\noindent \textbf{OVIS40:} OVIS40 is composed of visual instances of 40 categories of objects whose names are \textit{uncommon} nouns, \textit{e.g.}, ``\texttt{afro}'', ``\texttt{fresco}'', ``\texttt{pagoda}'' and are used as textual queries. In total, human labelers annotate 5,959 visual instances in 3,535 images for the 40 queries. On average, 149.0 visual instances are annotated for each query. None of the visual instances' name appears in the set of textual visual instance labels (seen labels) used for ILP during training. $88\%$ of the visual instances' names are not synonyms or hypernyms (super-classes) of any seen labels; $38\%$ of the names are hyponyms (sub-classes) of a seen label, $50\%$ of the names have no relation to any seen labels.

OVIS40 has three different subsets, \textit{i.e.}, OVIS40-small, OVIS40-medium, OVIS40-large. They differ in the numbers of distractors, \textit{i.e.}, images that do not contain any of the 40 categories of visual instances to be searched. The three subsets contain $\sim$10K, $\sim$20K and $\sim$114K distractors, respectively. The varied number of distractors ensures that the three subsets have different degree of difficulty.  OVIS40-large is particularly challenging as the number of distractors in it is 4$\times$, 10$\times$ and 32$\times$ more than those in OVIS-medium, OVIS-small and the number of images with annotated visual instances.

\noindent \textbf{OVIS1600:} OVIS1600 contains 1,600 different categories of visual instances, including visual instances of objects, motions and visual instances with certain attributes, which are to be searched using queries composed of nouns, verbs (\textit{e.g.}, ``\texttt{running}'', ``\texttt{standing}'') and adjectives (\textit{e.g.}, ``\texttt{equestrian}'', ``\texttt{misty}''). A total of 4,832 visual instances from 3,266 images are annotated. None of the queries in OVIS1600 appears in the set of textual visual instance labels (seen labels) used for ILP during training. $86\%$ of the 1,600 queries from OVIS1600 dataset are neither synonyms nor hypernyms (super-class) of any seen labels; $27\%$ of the queries are hyponyms (sub-class) of a seen label; $59\%$ have no relation to any seen labels. More importantly, $192$ queries ($12\%$) are adjectives, \textit{e.g.}, ``\texttt{equestrian}'', ``\texttt{misty}'' and $92$ queries ($6\%$) are verbs, while all seen labels are nouns.

There are a total of $\sim$117K images in OVIS1600, including $\sim$114K distractors (more than 95$\%$ of images are distractors). The large number of distractors not only simulate real application scenarios, but also make it quite challenging to find the visual instances to be searched. 

We refer our reader to the supplementary materials for more statistics about the two datasets.

\section{Experiments}

\subsection{Setup}

\noindent \textbf{Training Corpus.} We use three image captioning datasets, \textit{i.e.}, Conceptual Captions (CC) \cite{sharma2018conceptual}, SBU Captions \cite{ordonez2011im2text}  COCO Captions \cite{lin2014microsoft} to train our model (for MTP). CC is composed of $3.3$M image-caption pairs collected by a web crawler. SBU Captions and COCO Captions contain $870$K and $580$K image-caption pairs, respectively. We also use $98$K images with a set of $1,600$ categories of visual instance label annotations from VisualGenome \cite{krishna2017visual} (for ILP).

\renewcommand{\arraystretch}{1.0}
\begin{table*}[h]
    \centering
    \begin{tabular}{lccccccccc}
    \toprule
     & & \multicolumn{4}{c}{mAP} & \multicolumn{4}{c}{Precision} \\ 
    \cmidrule(lr){3-6} \cmidrule(lr){7-10}
     & Subset & mAP$_{30}$ & mAP$_{50}$ & mAP$_{70}$ & mAP$_\mathrm{all}$ & prec$_{30}$ & prec$_{50}$ & prec$_{70}$ & prec$_\mathrm{all}$ \\
    \midrule
    ILP & \multirow{3}{*}{OVIS40-medium} & 0.0 & 0.0 & 0.0 & 0.0 & 0.0 & 0.0 & 0.0 & 0.0 \\
    MTP & & 34.8 & 22.9 & 14.4 & 24.0 & 28.1 & 18.4 & 11.4 & 19.3\\
	MTP $\&$ ILP & & {\color{OrangeRed}46.9} & {\color{OrangeRed}30.6} & {\color{OrangeRed}14.4} &
    {\color{OrangeRed}30.6} & {\color{OrangeRed}47.0} & {\color{OrangeRed}31.5} & {\color{OrangeRed}16.6} & {\color{OrangeRed}31.6} \\
    \midrule
    ILP & \multirow{3}{*}{OVIS40-large} & 0.0 & 0.0 & 0.0 & 0.0 & 0.0 & 0.0 & 0.0 & 0.0 \\
    MTP & & 28.5 & 17.8 & {\color{OrangeRed}11.4} & 19.2 & 17.7 & 11.0 & 6.7 & 11.8 \\
    MTP $\&$ ILP & & {\color{OrangeRed}33.6} & {\color{OrangeRed}21.9} & 10.3 & {\color{OrangeRed}21.9} & {\color{OrangeRed}32.4} & {\color{OrangeRed}21.4} & {\color{OrangeRed}10.3} & {\color{OrangeRed}21.4} \\
    \bottomrule
    \end{tabular}
    \vspace{-2pt}
    \caption{Comparison of models trained using different training scheme. ILP: using visual instance label prediction only; MTP: using masked token prediction only; MTP $\&$ ILP: using both MTP and ILP (our proposed training scheme).}
    \label{tab:loss}
\end{table*}

\setlength{\tabcolsep}{3pt}
\renewcommand{\arraystretch}{1.0}
\begin{table*}[h]
    \centering
    \begin{tabular}{lcccccccccccc}
    \toprule
     & \multicolumn{4}{c}{OVIS40-small} & \multicolumn{4}{c}{OVIS40-medium} & \multicolumn{4}{c}{OVIS40-large} \\ 
    \cmidrule(lr){2-5} \cmidrule(lr){6-9} \cmidrule(lr){10-13}
    $\psi$ & mAP$_{30}$ & mAP$_{50}$ & mAP$_{70}$ & mAP$_\mathrm{all}$ & mAP$_{30}$ & mAP$_{50}$ & mAP$_{70}$ & mAP$_\mathrm{all}$ & mAP$_{30}$ & mAP$_{50}$ & mAP$_{70}$ & mAP$_\mathrm{all}$ \\ 
    \midrule
    cosine & 48.1 & 33.7 & 16.5 & 32.8 & 44.0 & 30.3 & {\color{OrangeRed}16.0} & 30.0 & 29.3 & 19.8 & 9.2 & 19.5 \\
    DP & {\color{OrangeRed}50.8} & {\color{OrangeRed}35.0} & {\color{OrangeRed}18.5} & {\color{OrangeRed}34.8} & {\color{OrangeRed}46.9} & {\color{OrangeRed}30.6} & 14.4 & {\color{OrangeRed}30.6} & {\color{OrangeRed}33.6} & {\color{OrangeRed}21.9} & {\color{OrangeRed}10.3} & {\color{OrangeRed}21.9} \\
    NDP & 49.1 & 32.8 & 17.7 & 33.2 & {\color{OrangeRed}46.9} & 30.2 & 14.6 & 30.5 & 31.7 & 19.1 & 8.3 & 19.7 \\
    \bottomrule
    \end{tabular}
    \vspace{-2pt}
    \caption{Comparison of the three choices for the similarity measure $\psi$ on OVIS40. cosine: cosine similarity; DP: dot product similarity; NDP: normalized dot product similarity. }
    \vspace{-10pt}
    \label{tab:phi}
\end{table*}
\setlength{\tabcolsep}{6pt}

\noindent \textbf{Implementation Details.} Our visual-semantic encoder is implemented as a $12$-layer Transformer encoder, with a hidden size of $768$. Its parameters are initialized with those of BERT-Base \cite{devlin2018bert}. The dictionary $D$ of our model contains $31,069$ tokens. Hence, the base-token embedding matrix $\mathbf{W}$ is of size $768 \times 31,069$. We train our ViSA model for 50 epochs with a batch size of 512 using AdamW optimizer \cite{loshchilov2017decoupled}. The learning rate is set to 0.00001.

\begin{figure*}[h]
    \centering
    \includegraphics[width=\linewidth]{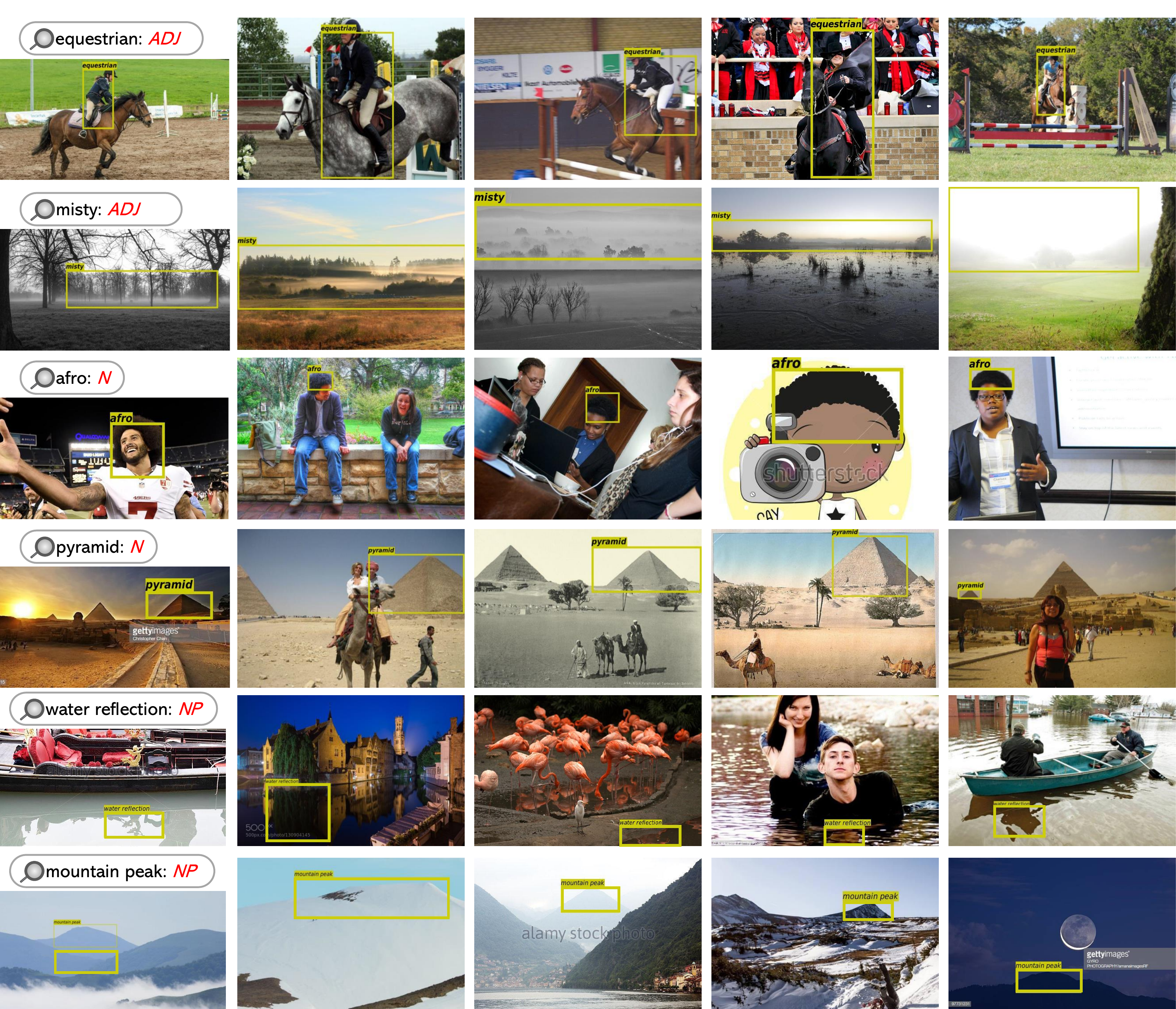}
    \caption{Visualization of top ranked visual instances returned by ViSA for six textual queries. Five queries have no relation to any visual instance labels used during training. ``\texttt{afro}'' is a hyponym of label ``hair''. {\color{red} \textit{ADJ, N, NP}} stand for adjective, noun and noun phrase, respectively.}
    \vspace{-10pt}
    \label{fig:pictures}
\end{figure*}

We adopt a Faster R-CNN, which is trained on VisualGenome using the \textit{same} visual instance label annotations we use for ILP (relationship between visual instances / queries in OVIS40 and OVIS1600 and the visual instance label annotations are discussed in Section \ref{sec:dataset}), to provide the positions of visual instances in images, and extract visual instance features with its ResNet101 \cite{he2016deep} backbone. While the Faster R-CNN is trained to detect $1,600$ categories objects, it performs surprisingly well at providing the positions of visual instances, even if the visual instances have no relation to any of the $1,600$ categories according to WordNet hierarchy \cite{wordnet} (as shown in Figure \ref{fig:pictures}). Note that traditional methods, \textit{e.g.}, EdgeBox \cite{zitnick2014edge} or Selective Search \cite{uijlings2013selective} can be used to \textit{directly} replace the Faster R-CNN to provide the positions of visual instances.

\noindent \textbf{Evaluation Metrics.} We evaluate the performance of OVIS method using mean average precision$@$k (mAP$@$k), which considers k top-ranked visual instances. We also adopt top-k precision (prec$@$k) as an auxiliary metric to show the percentage of true positives in the returned visual instances. We compute mAP and precision at three IoU thresholds: 30$\%$, 50$\%$ and $70\%$ and denote the results as mAP$@$k$_{30/50/70}$ and prec$@$k$_{30/50/70}$\footnote{``$@$k'' may be abbreviated if there is no confusion.}.

\subsection{Experiments on OVIS40}
\label{sec:ovis40-exp}

We adopt mAP$@$50 as the evaluation metric for all the experiments on OVIS40. The best performance is shown in {\color{OrangeRed}orange red} in all the tables.

\noindent \textbf{Comparison with DeViSE:} We compare the performances of our model, \textit{i.e.}, ViSA, and DeViSE\cite{frome2013devise} on all three subsets of OVIS40. As DeViSE is not able to perform OVIS, we modify it by combining it with a Faster R-CNN, which is the one used by our model, and adding all the queries in OVIS40 to DeViSE's dictionary so that it can learn their embeddings (this cannot be done in practice as queries are not known in advance). The modified model is trained using the same data as our model and is denoted as Det+DeViSE.

Table \ref{tab:ovis40-devise} shows the performance of ViSA and Det+DeViSE. We see that ViSA outperforms Det+DeViSE across all metrics. On OVIS-small, ViSA achieves 34.8$\%$ on mAP$_\all$, which is 5.4$\times$ more than that of Det+DeViSE. mAP$_\all$ of ViSA is 5.5$\times$ more than that of Det+DeViSE on OVIS-medium and is 6.3$\times$ more on OVIS-large. ViSA maintains it superiority over DeViSE as the number of distractors in the database rapidly increases. We also see that mAP$_\all$ of ViSA decreases from 34.8$\%$ to 30.4$\%$ and from 30.4$\%$ to 21.9$\%$, as the number of distractors increases by 10K (2$\times$) and by 94K (4.7$\times$). Despite that the number of distractors grows by \textit{9.4 times}, mAP$_\all$ of ViSA only decreases by 12.9$\%$, demonstrating ViSA's ability to handle tens of thousands of distractors and its potential to handle even larger number of distractors. 

\noindent \textbf{Comparison of Different Training Schemes:} To analyze our proposed training scheme, \textit{i.e.}, visual-semantic aligned (ViSA) representation learning, we conduct ablation studies by training our model using different components of ViSA. Table \ref{tab:loss} shows the performance of our model trained using different training schemes.

We can see from the 1$^\mathrm{st}$ row and the 4$^\mathrm{th}$ row that training with visual instance label prediction (ILP) results in a model that is not able to perform OVIS. The reason is that ILP only trains the model to predict textual labels of visual instances of a \textit{closed} set of categories. Thus, the trained model can not be used to search for other visual instances. If we train our model with masked token prediction (MTP) only (the 2$^\mathrm{nd}$ row and the 5$^\mathrm{th}$ row), the learned model achieves mAP$_\all$ of 24.0$\%$ and 19.2$\%$ on OVIS-medium and OVIS-large, respectively. This shows that our model implicitly learns to align representations of visual instances and textual search queries in a common semantic space with the help of MTP. The performance of our model becomes even better, if it is train with both MTP and ILP (our proposed training scheme). The increases in mAP$_\all$ and prec$_\all$ are 10.6$\%$ and 12.3$\%$ on OVIS40-medium and 2.7$\%$ and 9.6$\%$ on OVIS40-large. This MTP and ILP are \textit{complementary} to each other and are \textit{essential} for aligning the representations of visual instance and textual queries.

\noindent \textbf{Comparison of Similarity Measures} $\psi$\textbf{:} Table \ref{tab:phi} compares three different instantiations of the similarity measure $\phi$, \textit{i.e.}, cosine similarity, dot product similarity (DP) and normalized dot product similarity (NDP). Interestingly, they perform similarly. mAP$_\all$ of cosine, DP and NDP differ by less than 2.0$\%$, 0.6$\%$ and 2.4$\% $ on OVIS40-small, OVIS40-medium and OVIS40-large, respectively. The small gaps show that the performance of our model is not sensitive to the choice of the similarity measure $\psi$.

\begin{figure*}[h]
    \centering
        \includegraphics[trim={0 0 0 95pt}, clip, width=\linewidth]{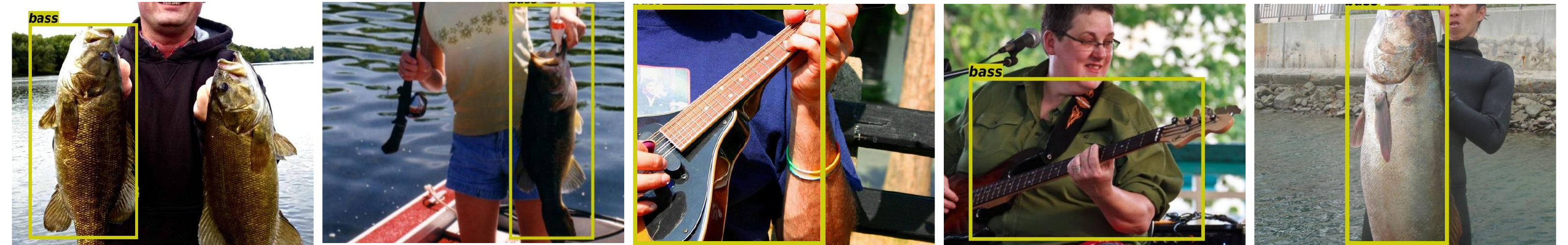}
        \captionsetup{singlelinecheck=false, justification=justified}
        \caption{Visualization of the visual instances returned by ViSA given a query ``\texttt{bass}''. ViSA returns visual instances that are relevant to \textit{both} of the two dramatically different meanings of ``\texttt{bass}''.
        \newline
        \texttt{bass}: (1) a name shared by many species of fish; (2) a member of the guitar family}
        \label{fig:bass}
        \vspace{-10pt}
\end{figure*}

\begin{figure*}
    \centering
    \includegraphics[trim={0 0 100pt 0}, clip, width=\linewidth]{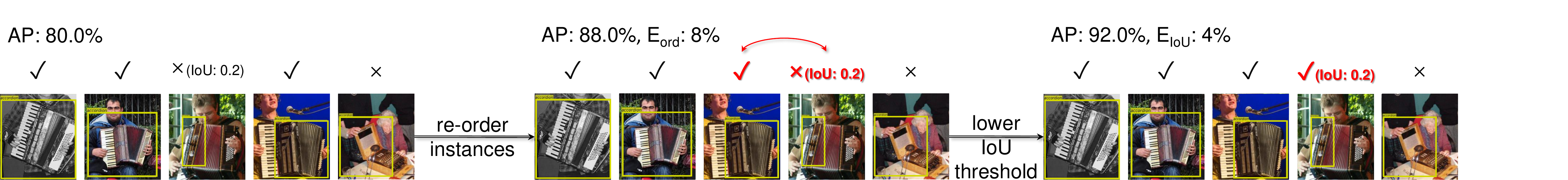}
    \vspace{-10pt}
    \caption{An illustration of the proposed error analysis pipeline. Given a search query ``\texttt{accordion}'', the original AP$@$5 is $80\%$. We first eliminate the order error by re-ordering the visual instances, such that TPs are ranked at higher orders than FPs. The change of AP before and after re-ordering is defined as the order error E$_\mathrm{ord}$ (8$\%$). We then eliminate the IoU error by lowering the IoU threshold to $0.01$. The increase in AP is defined as the IoU error E$_\mathrm{IoU}$ (4$\%$). The gap between the AP after lowering the IoU threshold and $100\%$ is defined as the background error E$_\mathrm{bg}$ (8$\%$).}
    \vspace{-3pt}
    \label{fig:error}
\end{figure*}

\subsection{Experiments on OVIS1600}

We adopt mAP$@$6 as the evaluation metric. Table \ref{tab:OVIS1100+-devise} shows a comparison of ViSA and DeViSE on OVIS1600 dataset. To enable DeViSE to perform OVIS, we make the same modifications to DeViSE as introduced in Section \ref{sec:ovis40-exp}. Comparing to Det+DeViSE, ViSA improves mAP by more than 10$\%$ across all IoU thresholds on OVIS1600. Specifically, ViSA achieves 12.4$\%$ on mAP$_\mathrm{70}$, which is computed at a rather high IoU threshold of 0.7, and also achieves 14.9$\%$ on mAP$_\mathrm{all}$. ViSA demonstrates its ability in searching for more than 1,600 queries in an image database composed of more than 117K images.

\renewcommand{\arraystretch}{1.0}
\begin{table}[h]
    \centering
\resizebox{1.\columnwidth}{!}{
    \begin{tabular}{lcccc}
    \toprule
    Model & mAP$_{30}$ & mAP$_{50}$ & mAP$_{70}$ & mAP$_\mathrm{all}$ \\
    \midrule
    Det+DeViSE \cite{frome2013devise} & 2.6 & 2.0 & 1.9 & 2.2 \\
    ViSA & {\color{OrangeRed}17.6} & {\color{OrangeRed}14.6} & {\color{OrangeRed}12.4} & {\color{OrangeRed}14.9} \\
    \bottomrule
    \end{tabular}
}
    \vspace{-6pt}
    \caption{Comparison of ViSA and a combination of a detector and DeViSE, \textit{i.e.}, Det+DeViSE, on OVIS1600.}
    \label{tab:OVIS1100+-devise}
\end{table}

\subsection{Qualitative Results}

Figure \ref{fig:pictures} shows the top-ranked visual instances returned by ViSA for six challenging queries from three different part of speeches, \textit{i.e.} adjective (``\texttt{equestrian}'', ``\texttt{misty}''), noun (``\texttt{afro}'', ``\texttt{pyramid}'') and noun phrase (``\texttt{water reflection}'', ``\texttt{mountain peak}''). We see that ViSA not only returns the images that contain the visual instances, but also accurately localize the visual instances.

Figure \ref{fig:bass} visualizes the top-ranked visual instances returned for a query ``\texttt{bass}'', which has two different meanings (1) a name shared by many species of fish, (2) a member of the guitar family. Interestingly, ViSA returns visual instances that are relevant to \textit{both} of the two dramatically different meanings of ``\texttt{bass}'', indicating that ViSA is capable of capturing \textit{multiple} aspects of the semantic meanings of a textual search query.

\subsection{Error Analysis}

We introduce a pipeline for analyzing errors made by OVIS methods, including but not limited to ViSA. There are three types of errors that prevent an OVIS method from achieving an mAP of $100\%$. (1) Order errors E$_\mathrm{ord}$ are caused by ranking false positives (FPs) at higher order than true positives (TPs). (2) IoU errors  E$_\mathrm{IoU}$ are caused by low IoU between the returned visual instances and the annotated visual instances. (3) Background errors are caused by returning visual instances from distractors, \textit{i.e.}, images that do not contain any visual instances relevant to the query. 

Figure \ref{fig:error} shows our proposed pipeline which quantitatively analyzes the influence of the three types of errors. The left most part of Figure \ref{fig:error} shows five top-ranked visual instances for query ``\texttt{accordion}''. We first eliminate the order error by re-ordering the list of returned visual instances, such that TPs are ranked at higher orders than FPs. The change of AP before and after re-ordering is defined as the order error E$_\mathrm{ord}$, which is 8$\%$ in this example. We then eliminate the IoU error by lowering the IoU threshold to $0.01$. The increase of AP brought by lowering the IoU threshold is defined as the IoU error E$_\mathrm{IoU}$, which is 4$\%$ in this example. The gap between the AP after lowering the IoU threshold and $100\%$ is defined as the background error E$_\mathrm{bg}$, which is 8$\%$ in this example. In the supplementary materials, we present an analysis of our model using such a pipeline.

\section{Conclusion}

In this work, we introduce the task of open-vocabulary visual instance search (OVIS), whose goal is to search for visual instances in a large-scale image database that are relevant to textual search queries. We propose a visual-semantic aligned representation learning (ViSA) method for OVIS. With the two complementary tasks of masked token prediction and visual instance label prediction, ViSA aligns representations of the visual instances and those of textual queries in a common semantic space, in which their similarities can be measured. We create two datasets, \textit{i.e.}, OVIS40 and OVIS1600, to benchmark OVIS methods, including ViSA. Experiments on both datasets verify the effectiveness of ViSA in performing OVIS.

{\small
\bibliographystyle{ieee_fullname}
\bibliography{egbib}

\begin{thebibliography}{10}\itemsep=-1pt

\bibitem{google}
\url{https://www.google.com/}.

\bibitem{ovis}
\url{https://en.wikipedia.org/wiki/Ovis}.

\bibitem{wordnet}
\url{https://wordnet.princeton.edu/}.

\bibitem{bentley1975multidimensional}
Jon~Louis Bentley.
\newblock Multidimensional binary search trees used for associative searching.
\newblock {\em Communications of the ACM}, 18(9):509--517, 1975.

\bibitem{berchtold1998fast}
Stefan Berchtold, Bernhard Ertl, Daniel~A Keim, H-P Kriegel, and Thomas Seidl.
\newblock Fast nearest neighbor search in high-dimensional space.
\newblock In {\em Proceedings 14th International Conference on Data
  Engineering}, pages 209--218. IEEE, 1998.

\bibitem{bilen2016weakly}
Hakan Bilen and Andrea Vedaldi.
\newblock Weakly supervised deep detection networks.
\newblock In {\em CVPR}, pages 2846--2854, 2016.

\bibitem{cao2020behind}
Jize Cao, Zhe Gan, Yu Cheng, Licheng Yu, Yen-Chun Chen, and Jingjing Liu.
\newblock Behind the scene: Revealing the secrets of pre-trained
  vision-and-language models.
\newblock {\em arXiv preprint arXiv:2005.07310}, 2020.

\bibitem{cao2018deep}
Yue Cao, Mingsheng Long, Bin Liu, and Jianmin Wang.
\newblock Deep cauchy hashing for hamming space retrieval.
\newblock In {\em CVPR}, pages 1229--1237, 2018.

\bibitem{chen2020uniter}
Yen-Chun Chen, Linjie Li, Licheng Yu, Ahmed El~Kholy, Faisal Ahmed, Zhe Gan, Yu
  Cheng, and Jingjing Liu.
\newblock Uniter: Universal image-text representation learning.
\newblock In {\em ECCV}, pages 104--120. Springer, 2020.

\bibitem{choe2020evaluating}
Junsuk Choe, Seong~Joon Oh, Seungho Lee, Sanghyuk Chun, Zeynep Akata, and
  Hyunjung Shim.
\newblock Evaluating weakly supervised object localization methods right.
\newblock In {\em CVPR}, pages 3133--3142, 2020.

\bibitem{devlin2018bert}
Jacob Devlin, Ming-Wei Chang, Kenton Lee, and Kristina Toutanova.
\newblock Bert: Pre-training of deep bidirectional transformers for language
  understanding.
\newblock {\em arXiv preprint arXiv:1810.04805}, 2018.

\bibitem{faghri2017vse++}
Fartash Faghri, David~J Fleet, Jamie~Ryan Kiros, and Sanja Fidler.
\newblock Vse++: Improving visual-semantic embeddings with hard negatives.
\newblock {\em arXiv preprint arXiv:1707.05612}, 2017.

\bibitem{frome2013devise}
Andrea Frome, Greg~S Corrado, Jon Shlens, Samy Bengio, Jeff Dean, Marc'Aurelio
  Ranzato, and Tomas Mikolov.
\newblock Devise: A deep visual-semantic embedding model.
\newblock In {\em Advances in neural information processing systems}, pages
  2121--2129, 2013.

\bibitem{gupta2019lvis}
Agrim Gupta, Piotr Dollar, and Ross Girshick.
\newblock Lvis: A dataset for large vocabulary instance segmentation.
\newblock In {\em Proceedings of the IEEE Conference on Computer Vision and
  Pattern Recognition}, pages 5356--5364, 2019.

\bibitem{he2016deep}
Kaiming He, Xiangyu Zhang, Shaoqing Ren, and Jian Sun.
\newblock Deep residual learning for image recognition.
\newblock In {\em CVPR}, pages 770--778, 2016.

\bibitem{hong2019asymmetric}
Weixiang Hong, Xueyan Tang, Jingjing Meng, and Junsong Yuan.
\newblock Asymmetric mapping quantization for nearest neighbor search.
\newblock {\em IEEE Transactions on Pattern Analysis and Machine Intelligence},
  2019.

\bibitem{hong2017fried}
Weixiang Hong, Junsong Yuan, and Sreyasee Das~Bhattacharjee.
\newblock Fried binary embedding for high-dimensional visual features.
\newblock In {\em CVPR}, pages 2749--2757, 2017.

\bibitem{huang2020comprehensive}
Zeyi Huang, Yang Zou, Vijayakumar Bhagavatula, and Dong Huang.
\newblock Comprehensive attention self-distillation for weakly-supervised
  object detection.
\newblock {\em Advances in neural information processing systems}, 2020.

\bibitem{hwang2012fast}
Yoonho Hwang, Bohyung Han, and Hee-Kap Ahn.
\newblock A fast nearest neighbor search algorithm by nonlinear embedding.
\newblock In {\em 2012 IEEE Conference on Computer Vision and Pattern
  Recognition}, pages 3053--3060. IEEE, 2012.

\bibitem{johnson2015image}
Justin Johnson, Ranjay Krishna, Michael Stark, Li-Jia Li, David Shamma, Michael
  Bernstein, and Li Fei-Fei.
\newblock Image retrieval using scene graphs.
\newblock In {\em Proceedings of the IEEE Conference on Computer Vision and
  Pattern Recognition}, pages 3668--3678, 2015.

\bibitem{krishna2017visual}
Ranjay Krishna, Yuke Zhu, Oliver Groth, Justin Johnson, Kenji Hata, Joshua
  Kravitz, Stephanie Chen, Yannis Kalantidis, Li-Jia Li, David~A Shamma, et~al.
\newblock Visual genome: Connecting language and vision using crowdsourced
  dense image annotations.
\newblock {\em International journal of computer vision}, 123(1):32--73, 2017.

\bibitem{li2020hero}
Linjie Li, Yen-Chun Chen, Yu Cheng, Zhe Gan, Licheng Yu, and Jingjing Liu.
\newblock Hero: Hierarchical encoder for video+ language omni-representation
  pre-training.
\newblock In {\em EMNLP}, 2020.

\bibitem{li2020weakly}
Liunian~Harold Li, Haoxuan You, Zhecan Wang, Alireza Zareian, Shih-Fu Chang,
  and Kai-Wei Chang.
\newblock Weakly-supervised visualbert: Pre-training without parallel images
  and captions.
\newblock {\em arXiv preprint arXiv:2010.12831}, 2020.

\bibitem{li2020oscar}
Xiujun Li, Xi Yin, Chunyuan Li, Pengchuan Zhang, Xiaowei Hu, Lei Zhang, Lijuan
  Wang, Houdong Hu, Li Dong, Furu Wei, et~al.
\newblock Oscar: Object-semantics aligned pre-training for vision-language
  tasks.
\newblock In {\em ECCV}, pages 121--137. Springer, 2020.

\bibitem{li2016adaptive}
Zhujin Li, Xianglong Liu, Junjie Wu, and Hao Su.
\newblock Adaptive binary quantization for fast nearest neighbor search.
\newblock In {\em ECAI}, pages 64--72, 2016.

\bibitem{lin2014microsoft}
Tsung-Yi Lin, Michael Maire, Serge Belongie, James Hays, Pietro Perona, Deva
  Ramanan, Piotr Doll{\'a}r, and C~Lawrence Zitnick.
\newblock Microsoft coco: Common objects in context.
\newblock In {\em ECCV}, pages 740--755. Springer, 2014.

\bibitem{liu2016deep}
Haomiao Liu, Ruiping Wang, Shiguang Shan, and Xilin Chen.
\newblock Deep supervised hashing for fast image retrieval.
\newblock In {\em CVPR}, pages 2064--2072, 2016.

\bibitem{loshchilov2017decoupled}
Ilya Loshchilov and Frank Hutter.
\newblock Decoupled weight decay regularization.
\newblock {\em ICLR}, 2019.

\bibitem{oquab2015object}
Maxime Oquab, L{\'e}on Bottou, Ivan Laptev, and Josef Sivic.
\newblock Is object localization for free?-weakly-supervised learning with
  convolutional neural networks.
\newblock In {\em CVPR}, pages 685--694, 2015.

\bibitem{ordonez2011im2text}
Vicente Ordonez, Girish Kulkarni, and Tamara Berg.
\newblock Im2text: Describing images using 1 million captioned photographs.
\newblock {\em Advances in neural information processing systems},
  24:1143--1151, 2011.

\bibitem{ren2020instance}
Zhongzheng Ren, Zhiding Yu, Xiaodong Yang, Ming-Yu Liu, Yong~Jae Lee,
  Alexander~G Schwing, and Jan Kautz.
\newblock Instance-aware, context-focused, and memory-efficient weakly
  supervised object detection.
\newblock In {\em Proceedings of the IEEE/CVF Conference on Computer Vision and
  Pattern Recognition}, pages 10598--10607, 2020.

\bibitem{sharma2018conceptual}
Piyush Sharma, Nan Ding, Sebastian Goodman, and Radu Soricut.
\newblock Conceptual captions: A cleaned, hypernymed, image alt-text dataset
  for automatic image captioning.
\newblock In {\em Proceedings of the 56th Annual Meeting of the Association for
  Computational Linguistics}, pages 2556--2565, 2018.

\bibitem{spink2001searching}
Amanda Spink, Dietmar Wolfram, Major~BJ Jansen, and Tefko Saracevic.
\newblock Searching the web: The public and their queries.
\newblock {\em Journal of the American society for information science and
  technology}, 52(3):226--234, 2001.

\bibitem{su2019vl}
Weijie Su, Xizhou Zhu, Yue Cao, Bin Li, Lewei Lu, Furu Wei, and Jifeng Dai.
\newblock Vl-bert: Pre-training of generic visual-linguistic representations.
\newblock {\em ICLR}, 2020.

\bibitem{tan2019lxmert}
Hao Tan and Mohit Bansal.
\newblock Lxmert: Learning cross-modality encoder representations from
  transformers.
\newblock {\em arXiv preprint arXiv:1908.07490}, 2019.

\bibitem{uijlings2013selective}
Jasper~RR Uijlings, Koen~EA Van De~Sande, Theo Gevers, and Arnold~WM Smeulders.
\newblock Selective search for object recognition.
\newblock {\em International journal of computer vision}, 104(2):154--171,
  2013.

\bibitem{vaswani2017attention}
Ashish Vaswani, Noam Shazeer, Niki Parmar, Jakob Uszkoreit, Llion Jones,
  Aidan~N Gomez, {\L}ukasz Kaiser, and Illia Polosukhin.
\newblock Attention is all you need.
\newblock In {\em Advances in neural information processing systems}, pages
  5998--6008, 2017.

\bibitem{wen2016discriminative}
Yandong Wen, Kaipeng Zhang, Zhifeng Li, and Yu Qiao.
\newblock A discriminative feature learning approach for deep face recognition.
\newblock In {\em ECCV}, pages 499--515. Springer, 2016.

\bibitem{wu2020forest}
Jialian Wu, Liangchen Song, Tiancai Wang, Qian Zhang, and Junsong Yuan.
\newblock Forest r-cnn: Large-vocabulary long-tailed object detection and
  instance segmentation.
\newblock In {\em Proceedings of the 28th ACM International Conference on
  Multimedia}, pages 1570--1578, 2020.

\bibitem{yang2019towards}
Ke Yang, Dongsheng Li, and Yong Dou.
\newblock Towards precise end-to-end weakly supervised object detection
  network.
\newblock In {\em ICCV}, pages 8372--8381, 2019.

\bibitem{ye2019cap2det}
Keren Ye, Mingda Zhang, Adriana Kovashka, Wei Li, Danfeng Qin, and Jesse
  Berent.
\newblock Cap2det: Learning to amplify weak caption supervision for object
  detection.
\newblock In {\em ICCV}, pages 9686--9695, 2019.

\bibitem{yu2020product}
Tan Yu, Jingjing Meng, Chen Fang, Hailin Jin, and Junsong Yuan.
\newblock Product quantization network for fast visual search.
\newblock {\em International Journal of Computer Vision}, pages 1--19, 2020.

\bibitem{yu2017hope}
Tan Yu, Yuwei Wu, and Junsong Yuan.
\newblock Hope: Hierarchical object prototype encoding for efficient object
  instance search in videos.
\newblock In {\em CVPR}, pages 2424--2433, 2017.

\bibitem{yu2018product}
Tan Yu, Junsong Yuan, Chen Fang, and Hailin Jin.
\newblock Product quantization network for fast image retrieval.
\newblock In {\em ECCV}, pages 186--201, 2018.

\bibitem{zeng2019wsod2}
Zhaoyang Zeng, Bei Liu, Jianlong Fu, Hongyang Chao, and Lei Zhang.
\newblock Wsod2: Learning bottom-up and top-down objectness distillation for
  weakly-supervised object detection.
\newblock In {\em ICCV}, pages 8292--8300, 2019.

\bibitem{zhang2020context}
Qi Zhang, Zhen Lei, Zhaoxiang Zhang, and Stan~Z Li.
\newblock Context-aware attention network for image-text retrieval.
\newblock In {\em Proceedings of the IEEE/CVF Conference on Computer Vision and
  Pattern Recognition}, pages 3536--3545, 2020.

\bibitem{zhou2016learning}
Bolei Zhou, Aditya Khosla, Agata Lapedriza, Aude Oliva, and Antonio Torralba.
\newblock Learning deep features for discriminative localization.
\newblock In {\em CVPR}, pages 2921--2929, 2016.

\bibitem{zhou2020unified}
Luowei Zhou, Hamid Palangi, Lei Zhang, Houdong Hu, Jason~J Corso, and Jianfeng
  Gao.
\newblock Unified vision-language pre-training for image captioning and vqa.
\newblock In {\em AAAI}, pages 13041--13049, 2020.

\bibitem{zitnick2014edge}
C~Lawrence Zitnick and Piotr Doll{\'a}r.
\newblock Edge boxes: Locating object proposals from edges.
\newblock In {\em European conference on computer vision}, pages 391--405.
  Springer, 2014.

\end{thebibliography}
}

\end{document}